\documentclass[10pt,twocolumn,letterpaper]{article}

\usepackage[pagenumbers]{wacv} 

\usepackage{graphicx}
\usepackage{amsmath}
\usepackage{amssymb}
\usepackage{booktabs}
\usepackage{times}
\usepackage{epsfig}
\usepackage[accsupp]{axessibility}  
\usepackage{listings}
\usepackage{xcolor}
\lstdefinestyle{jupyterpython}{
  language=Python,
  backgroundcolor=\color{gray!10}, 
  basicstyle=\ttfamily\footnotesize, 
  keywordstyle=\bfseries\color{blue!90}, 
  commentstyle=\color{green!60!black}, 
  stringstyle=\color{red!70!black}, 
  frame=single, 
  breaklines=true, 
  showstringspaces=false, 
  tabsize=4 
}

\usepackage{multirow}
\usepackage{graphicx}

\usepackage[pagebackref,breaklinks,colorlinks]{hyperref}

\usepackage[capitalize]{cleveref}
\crefname{section}{Sec.}{Secs.}
\Crefname{section}{Section}{Sections}
\Crefname{table}{Table}{Tables}
\crefname{table}{Tab.}{Tabs.}

\title{Detecting Contextual Anomalies  by Discovering Consistent Spatial Regions}

\author{Zhengye Yang\\
Rensselaer Polytechnic Institute\\
Department of ECSE, Troy, NY, USA\\
{\tt\small yangz15@rpi.edu}
\and
Richard J.~Radke\\
Rensselaer Polytechnic Institute\\
Department of ECSE, Troy, NY, USA\\
{\tt\small rjradke@ecse.rpi.edu}
}

\def\wacvPaperID{10} 
\def\confName{WACV}
\def\confYear{2025}

\newcommand{\vf}{{\mathbf{f}}}

\newcommand{\vl}{{\mathbf{l}}}

\newcommand{\vv}{{\mathbf{v}}}

\newcommand{\vx}{{\mathbf{x}}}

\newcommand{\vM}{{\mathbf{M}}}

\newcommand{\vO}{{\mathbf{O}}}

\newcommand{\RR}{\mathbb{R}}

\newcommand{\vone}{{\mathbf{1}}}

\newcommand{\rnote}[1]{\textcolor{red}{#1}}

\begin{document}
\maketitle
\begin{abstract}
We describe a method for modeling spatial context to enable video anomaly detection. The main idea is to discover regions that share similar object-level activities by clustering joint object attributes using Gaussian mixture models.  We demonstrate that this straightforward approach, using orders of magnitude fewer parameters than competing models, achieves state-of-the-art performance in the challenging spatial-context-dependent Street Scene dataset.  As a side benefit,  the high-resolution discovered regions learned by the model also provide explainable normalcy maps for human operators without the need for any pre-trained segmentation model.
\end{abstract}    
\section{Introduction}
\label{sec:intro}

Determining whether an event observed in a video stream is anomalous often depends on its spatial \emph{context}. For example, pedestrians are expected on sidewalks and cyclists in bike lanes, but not vice versa. In real-world video anomaly detection, the appearance and motion of objects alone are insufficient to classify anomalies; we should learn the “normal” behavior of objects at different locations in the video based on long-term observation.

A major trend in modern video anomaly detection (VAD) algorithms is to train convolutional auto-encoders using a reconstruction proxy task \cite{hasan_learning_2016,gong_memorizing_2019,park_learning_2020}. Future frame prediction \cite{liu_future_2018,park_learning_2020,lv_learning_2021,liu_diversity-measurable_2023} can be considered a special form of reconstruction that also addresses temporal regularity. These reconstruction-based methods have achieved great success on widely adopted VAD benchmark datasets, such as UCSD Ped2 \cite{weixin_li_anomaly_2014}, CUHK Avenue \cite{lu_abnormal_2013}, and ShanghaiTech \cite{liu_future_2018}. However, we observe a large performance drop when such algorithms are applied to datasets in which spatial context plays a critical role in making the anomaly decision, such as the relatively new Street Scene benchmark dataset  \cite{ramachandra_street_2020}. One reason is that convolutional auto-encoders are poorly suited to learning spatial context due to their shift-invariant nature \cite{goyal_inductive_2022}.

\begin{figure}[!htbp]
    \centering
    \includegraphics[width=1\linewidth]{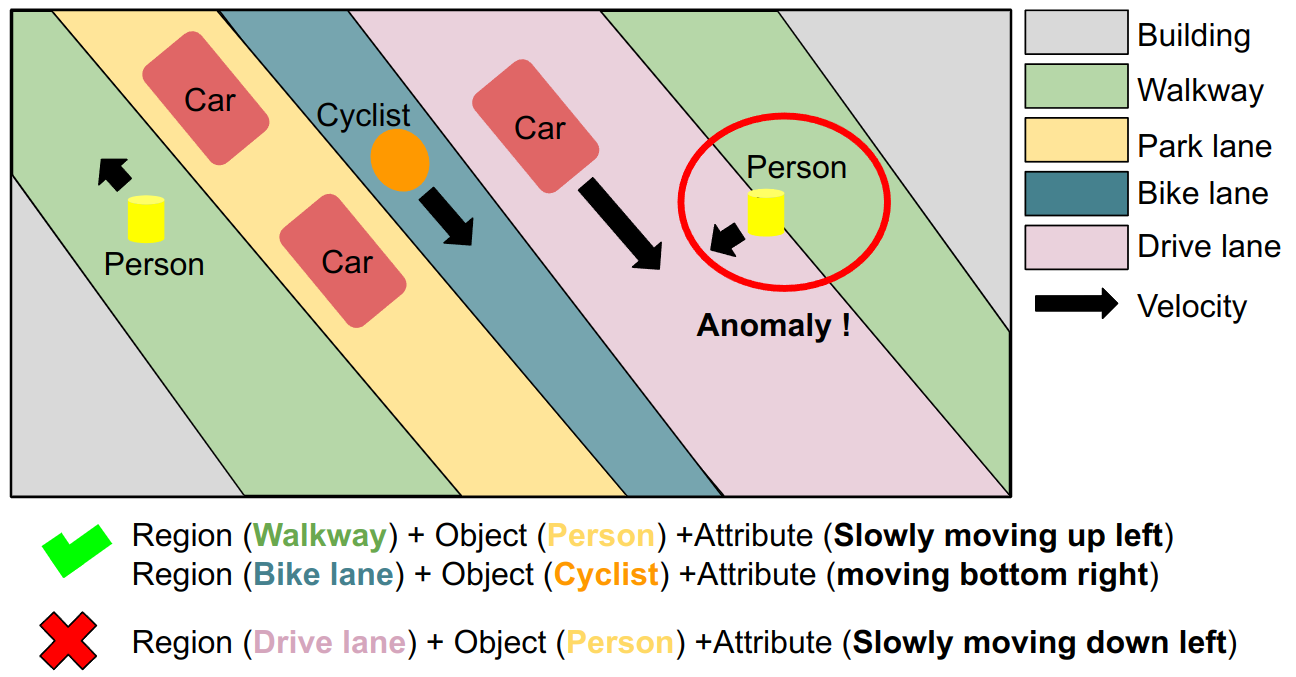}
    \caption{A toy example of building spatial context. Using the discovered semantically meaningful regions, spatial-context-dependent anomalies can be easily detected.}
    \label{fig:spatial_toy}
    \vspace{-0.4cm}
\end{figure}

\begin{figure*}[htbp!]
\begin{center}
\includegraphics[width=0.99\linewidth]{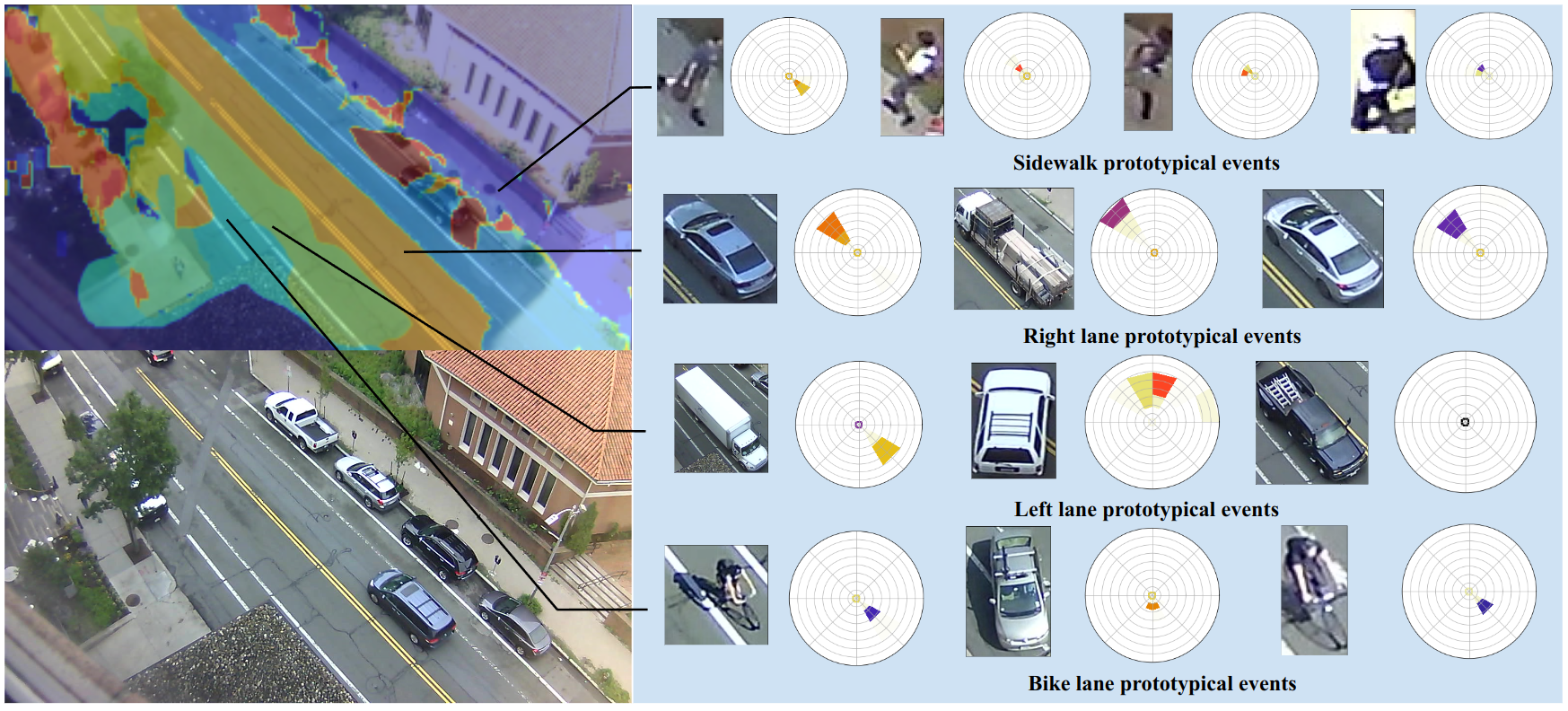}
\end{center}
   \caption{The proposed method discovers regions that have similar object and motion (top left image), illustrated here in the Street Scene dataset (lower left image) \cite{ramachandra_street_2020}.  Each region is characterized by a learned mixture of prototypical events in appearance/motion feature space.  On the right side we show motion descriptors in magnitude/angle space for learned modes of the mixture model in several regions, along with corresponding objects. }
\label{fig:semantic_nn}
\vspace{-0.4cm}
\end{figure*} 

In contrast, high-performing VAD algorithms on the Street Scene dataset (e.g., \cite{singh_eval_2022,ramachandra_street_2020}) usually separate videos into hundreds of rectangular subregions, each with a dedicated normalcy model. Hundreds of subregions not only result in hundreds of normalcy models but also reduce model robustness due to the small number of available object samples per region. However, a typical surveillance camera does not capture hundreds of small regions with different normalcy patterns; meaningful regions should be automatically discovered from long-term observations.

Fig.~\ref{fig:spatial_toy} illustrates a toy experiment mimicking a typical street scenario. Suppose we can extract representations like “pedestrians walk or jog on the sidewalk,” “cyclists ride within the bike lane,” and “cars move quickly in the traffic lane in the correct direction” from training videos. Then, given a new event represented in the same “region-object-attribute” form, we should be able to achieve an almost perfect anomaly detection result. Leveraging object categories and their corresponding attributes is already widespread in the VAD community \cite{micorek2024mulde,reiss_attribute-based_2022,sun_hierarchical_2023}. However, automatically extracting consistent spatial regions based on these behaviors and using them as the basis for anomaly detection is relatively unexplored territory.

In this paper, we describe a natural approach to discovering spatial regions that have similar expected object motion and appearance, and make anomaly decisions based on what is “normal” for a given region. With high-quality region discovery, we show that training a simple Gaussian mixture model suffices to detect spatial-context-dependent anomalies. This has the side benefits of greatly reducing the number of models required for learning and inference, as well as providing better interpretability to the user.


Our approach achieves state-of-the-art performance on the Street Scene dataset using straightforward Gaussian mixture models to discover a small number of activity regions specified by the user, each defined by a set of prototypical events (modes) in a low-dimensional appearance/motion space. As illustrated in Fig.~\ref{fig:semantic_nn}, we automatically discover semantically meaningful regions like traffic lanes, bike lanes, and sidewalks, each of which has typical object appearance and motion. For a new object, we can classify its track as normal or anomalous based on its likelihood according to the learned region model, discovering spatial-context-dependent anomalies like pedestrians in the street or illegally parked cars. The contributions of this paper are threefold:\
\begin{itemize}
\item We propose a pixel-level clustering method for region discovery and compare it against the common approach of rectangular patches, validating the effectiveness of our strategy in terms of both performance and efficiency.
\item We propose a highly effective approach focusing on spatial-context anomalies that achieves state-of-the-art VAD performance on the Street Scene dataset.
\item We demonstrate that the discovered regions provide natural interpretations of the expected normal activities within each region.
\end{itemize}

\section{Related Work}
\label{sec:related}
\begin{figure}[htbp!]
       \centering
       \includegraphics[width=1\linewidth]{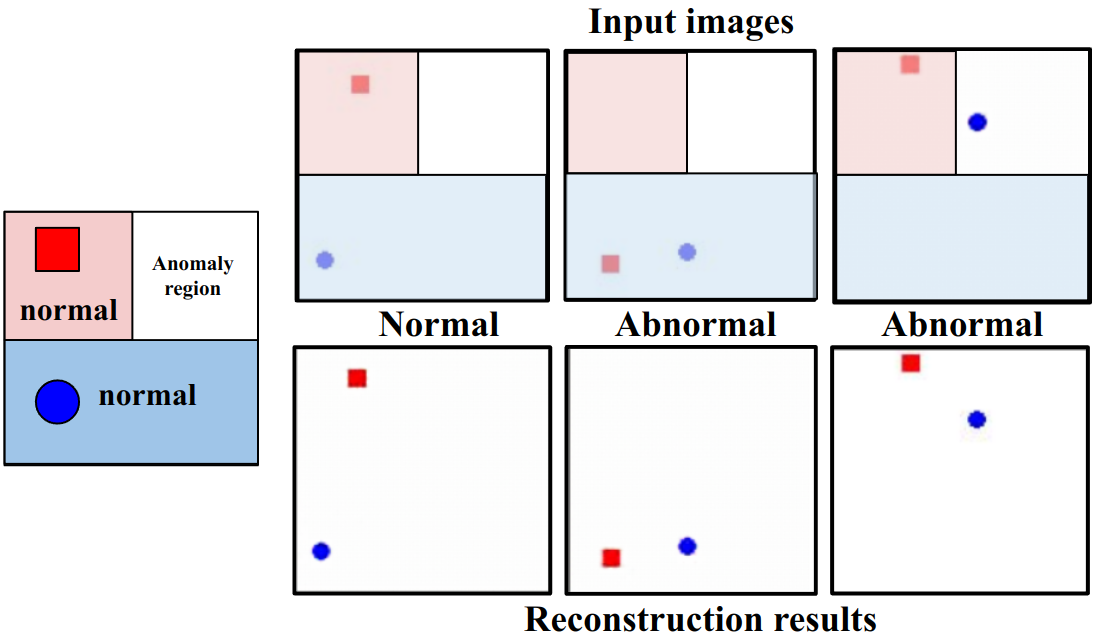}
       \caption{A CNN-based autoencoder fails to learn spatial context due to its shift-invariant nature. The autoencoder can perfectly reconstruct both normal and abnormal samples without understanding what is allowable in each region.}
       \label{fig:cnn_inductive}
       \vspace{-0.2cm}
   \end{figure}

\noindent\textbf{Reconstruction-based Methods.}
Early VAD approaches were based on dictionary learning (e.g., \cite{lu_abnormal_2013}). The successors to this approach are modern reconstruction-based methods \cite{hasan_learning_2016,gong_memorizing_2019,lv_learning_2021,park_learning_2020,yang_video_2023,cho_unsupervised_2022,zhang_multi-scale_2024} and future-frame-prediction-based methods \cite{liu_future_2018, liu_diversity-measurable_2023,park_learning_2020,lv_learning_2021}. In both types of approaches, an auto-encoder is trained to perform reconstruction or prediction, and anomalies are detected based on the reconstruction error of query frames. However, we observe that convolutional neural networks (CNNs), the building blocks shared by many auto-encoders, are naturally unsuitable for learning spatial context due to their shift-invariant properties. Figure \ref{fig:cnn_inductive} illustrates a toy experiment to demonstrate this issue, using the same normal/anomalous classification setup used by most VAD methods. The training data is generated so that red squares only appear in the upper left quadrant and blue circles only appear in the bottom half. Half of the testing data is generated following the same rule, while the other half generates the square and circle positions freely. We see that a convolutional auto-encoder trained from the normal data can also perfectly reconstruct the anomalous frames; spatial context has not been learned. Thus, these types of methods typically perform poorly on real datasets like Street Scene that contain spatial-context-dependent anomalies.

\noindent\textbf{Object-centric Methods.}
Methods that use pre-trained networks to extract highly discriminative object-centric features often demonstrate superior performance on traditional benchmark datasets \cite{liu_hybrid_2021,ionescu_object-centric_2019,georgescu_background-agnostic_2021,sun_hierarchical_2023,hirschorn_normalizing_2023}. However, their ability to learn spatial context is relatively limited since many benchmark datasets contain only a few anomalous examples that depend on spatial context.

\noindent\textbf{Context Learning in VAD.}
As VAD algorithms’ performance improves, the state of the art on several benchmarks is basically saturated (e.g., over 99.0 AUC on the UCSD Ped2 dataset). Recent research has pivoted to more nuanced normalcy modeling to detect context-dependent anomalies. Bao \etal \cite{bao_hierarchical_2022} proposed pixel-level clustering to generate pseudo-labels to train a background segmentation model to help an object-centric auto-encoder predict the next frame. Sun \etal \cite{sun_hierarchical_2023} proposed a pre-trained segmentation model to create scene-level background features concatenated with motion and appearance vectors to perform scene-aware VAD. However, we must emphasize that spatial context should not be reduced to background appearance (for example, opposing lanes of traffic may have the same background appearance). Further, scene dependence only constrains the normality to be associated with specific cameras or scenes instead of regions within a camera. It is also important to decouple background appearance and spatial context to develop unbiased VAD algorithms.

Yang and Radke \cite{yang_context-aware_2024} proposed a contrastive learning method to learn spatial-temporal context on the frame and patch level, focusing on determining temporal context-dependent anomalies. Singh \etal’s EVAL \cite{singh_eval_2022} followed a similar strategy to the original Street Scene paper \cite{ramachandra_street_2020}, separating a video clip into hundreds of overlapping spatial tubes and storing region-specific exemplar features extracted from a set of attribute networks to detect spatially dependent anomalies. The follow-on method T-EVAL \cite{singh_tracklet-based_2024} stores tracklets with corresponding visual features as exemplars and uses nearest neighbor search to detect anomalies. It is worth noting that the search is performed within a 3x3 grid region centered on the query exemplar location to speed up the search process. This can still be considered grid-based modeling. Although there is no dedicated model for each region, when the number of training videos increases, both storage and computational complexity become unmanageable. Our method discovers spatial semantic regions that drastically decrease the number of dedicated models required to cover the scene and achieves better VAD performance compared to these methods.

\noindent\textbf{Explainability in VAD.}
Explainability in VAD is a desirable property addressed by prior work in various ways \cite{wu_explainable_2022,singh_tracklet-based_2024,szymanowicz_discrete_2022}. Szymanowicz \etal \cite{szymanowicz_discrete_2022} used a pre-trained action recognition module to provide action labels for each detected anomaly bounding box as a way to explain the anomalous events. Wu \etal and EVAL \cite{singh_tracklet-based_2024} both explain anomalies using high-level attribute features. Recently, several approaches have adopted Large Language Models (LLMs) and Multimodal Large Language Models (MLLMs) for the VAD task \cite{zhang_holmes-vad_2024,lv_video_2024,bharadwaj_vane-bench_2024}. While both LLMs and MLLMs show promise in reasoning and explainability, their success in VAD tasks is due more to the huge overlap between commonsense knowledge encoded in those models and the definition of abnormality in benchmark datasets. Moreover, such commonsense knowledge might introduce harmful bias in determining anomalies using appearance, background bias, or lack of context for the specific area that the camera covers. Our proposed method can provide statistical patterns of appearance and motion that could be further passed to LLMs or MLLMs for natural-language-based explainable VAD.

\noindent\textbf{Region discovery.}
An older line of work investigates region discovery strategies in video \cite{chen_change_loy_multi-camera_2009,fleet_finding_2014,wang_detecting_2020,zhou_measuring_2014}. For example, \cite{chen_change_loy_multi-camera_2009} uses a pixel-level foreground ratio within a time window to find similar activity regions through correlation. Wang \etal \cite{fleet_finding_2014} uses a thermal-diffusion based approach on the optical flow field with clustering to create semantic regions. However, this region clustering is performed at the pixel level and can mainly handle collectively moving dense particles or large objects with a unified motion. Our region discovery method is based on object-level activity and has no requirement about objects' collective motion or size.   
\begin{figure*}
\begin{center}
\includegraphics[width=0.99\linewidth]{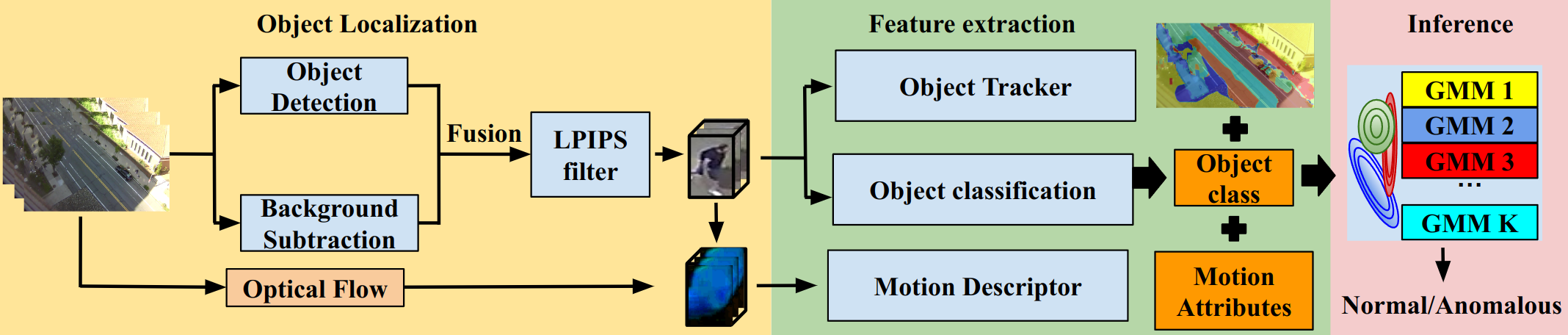}
\end{center}
   \caption{Overview of the proposed system inference pipeline. Extracted object level features are assigned to different GMM models based on region discovery results. }
\label{fig:system_overview}
\vspace{-0.4cm}
\end{figure*}  
\section{Method}
\label{sec:method}
As stated in the thought experiment, our goal is to build a model that approximates the ``region-object-attribute" idea. To achieve this, our method has three key stages: object representation, region discovery, and anomaly detection, as illustrated in Figures \ref{fig:system_overview} and \ref{fig:fullresupdate}. We rely on the object representation to discover semantically meaningful regions and to train normalcy models. To define the problem, suppose we have a collection of $N$ training video clips from the same stationary camera, $\{\vv_{1},...,\vv_{N}\} \in \RR^{T \times H \times W \times C} $, where $T, H, W, C$ denote the temporal window size, height, width and color channel respectively. We will create a $H \times W$ map corresponding to $K$ spatial regions, each of which contains a similar set of activities.  The learned model in each region then forms the basis for spatial-context-dependent anomaly detection.

\subsection{Object representation}

Many object-centric VAD methods \cite{georgescu_background-agnostic_2021,bao_hierarchical_2022,liu_hybrid_2021,sun_hierarchical_2023,reiss_attribute-based_2022} use pretrained networks to extract various visual features.  We also follow this approach, forming tracklets and extracting object attributes from detected object regions.

\noindent\textbf{Object region proposal.} We first feed each video clip $\vx$ into a pre-trained YOLOv8 model \cite{yolov8} to obtain object bounding boxes. Due to occlusions, varying lighting conditions, and the top-down perspective in the Street Scene dataset, solely relying on bounding boxes from object detection results in many missed detections. To capture as many bounding boxes as possible, we merge the detection results with foreground objects detected using a simple Gaussian mixture background model. 

\noindent\textbf{False region removal.} Using the fused detection strategy above provides excellent bounding box proposals that include most of the ground-truth objects. However, both object detection and foreground subtraction  introduce a significant number of region proposals that only cover pieces of background regions due to false detections. This problem can be greatly mitigated by using LPIPS \cite{zhang_unreasonable_2018} to compare each region proposal with the learned background, retaining proposals with LPIPS loss greater than a threshold $\alpha$. 

\noindent\textbf{Object class descriptor.} 
Although region proposals from object detection contain category information, region proposals from background subtraction do not.  
For each detected bounding box, we apply a pre-trained CLIP vision encoder (Vit-B32) \cite{radford_learning_2021}  and use the CLS token from its output as an appearance feature vector $\vf_{app} \in \RR^{512}$. We perform zero-shot classification with a set of pre-selected expected classes (person, bicycle, car, motorcycle) with the text prompt templates used for Imagenet classification in \cite{radford_learning_2021}. We use one-hot encoding to represent the object class information  $\vf_{obj} \in \RR^{4}$ as a compact and highly abstract representation for the following process. 

\noindent\textbf{Motion descriptor.} We compute the Farneback optical flow \cite{farneback2003two} of the whole clip $\vv_{flow} \in \RR^{(T-1) \times H \times W \times 2}$ to compute motion attributes. For each detected bounding box, we use Histogram-of-Flow (HOF) to summarize the motion $\vv_{flow}$ within the box.  We use 12 uniformly spaced orientation bins and 1 background ratio bin.  Pixels whose magnitude are below a certain threshold (1.5 in our experiment) are considered as background pixels and aggregated in a single bin. The average speed is calculated for each orientation bin. The motion attributes $\vf_{mot}$ for the training/inference normalcy model are the dominant orientation and corresponding speed of the given tracklet resulting in $\vf_{mot} \in \RR^{2}$.

\noindent\textbf{Tracklet formulation.} To incorporate temporal information, we use the DeepSORT algorithm \cite{Wojke2017simple} to associate bounding box identities across time with the extracted CLIP embedding $\vf_{app}$. Since the full trajectory lengths might vary, we create tracklets $\vl \in\RR^{2 \times t_{w} }$ with a fixed length window $t_{w}$. Thus for each detected object we formulate a combined feature $\vO^{i}_{t} = [\vf_{obj}^{i},\vf_{mot}^{i},\vl^{i}]_{t}$ at each time stamp $t$, which we use as the basis for the following process. 


\subsection{Region Discovery Strategies}

In frame-based methods, each spatial location has an equal amount of training data. This is not the case when considering the alternative object-centric route. The region representation for object-centric methods requires more careful design to both reduce the number of separate models and improve detection performance in regions where a limited number of samples is available. 


\noindent\textbf{Grid Region Separation.} The most straightforward way to introduce spatial context in object-centric VAD is to split the image into a non-overlapping rectangular grid, design a model for each rectangle, and make a frame-level determination about each object based on the rectangle containing the center point of its detected bounding box.

In our experiment, we found that using small patches (e.g., 40$\times$40 as used in \cite{ramachandra_street_2020}) results in a large number of patches that don't contain a single example; thus no model can be trained for those regions. We chose an 80$\times$80 rectangle size as our baseline in our comparisons below. This large patch setting still generates 144 separate models.  


\noindent\textbf{Region Discovery.}  Our key innovation is to cluster regions with similar appearance/motion patterns, resulting in a substantial decrease in the number of models to be trained.  Our clustering approach is agnostic to the type of feature representation. 

To create a high-resolution (around 1 million pixel) region map, we need to first discover what object-level events happen at each pixel location.  For each object tracklet moving through the pixel, we determine its dominant motion direction and quantize its speed into 4 log-scale-spaced magnitude bins. Thus for each object, we acquire its class information, dominant motion direction, and speed. We initialize an empty heatmap $\vM \in \RR^{H\times W \times D}$ with the same size as the original frame and $D$ channels corresponding to the given attributes, which we will incrementally update using the training data.


To provide enough information to create a high-resolution map, we center a Gaussian kernel around the tracklet pixel  $x_c,y_c$, using the $\sigma$ parameter to control the spatial influence of each moving object on surrounding pixels (and zeroing out the contribution outside the object bounding box). We then add this to the incrementally constructed heatmap $\vM$ in (Eqn.~\ref{eq:heatmap_update}).  

\begin{equation}
\begin{aligned}
\mathbf{G}(x, y) &= \exp\left( - \frac{(x - x_c)^2 + (y - y_c)^2}{2 \sigma^2} \right) \\
\vone_{\text{box}}(x, y) &= 
    \begin{cases} 
        1 & \text{if } x_{tl} \leq x \leq x_{br} \text{ and } y_{tl} \leq y \leq y_{br}, \\
        0 & \text{otherwise},
    \end{cases} \\
\mathbf{M}(x, y, d) &\gets \mathbf{M}(x, y, d) + \vone_{\text{box}}(x, y) \cdot \mathbf{G}(x, y) 
\end{aligned}
\label{eq:heatmap_update}
\end{equation}

\begin{figure}[htbp!]
    \centering
    \includegraphics[width=1\linewidth]{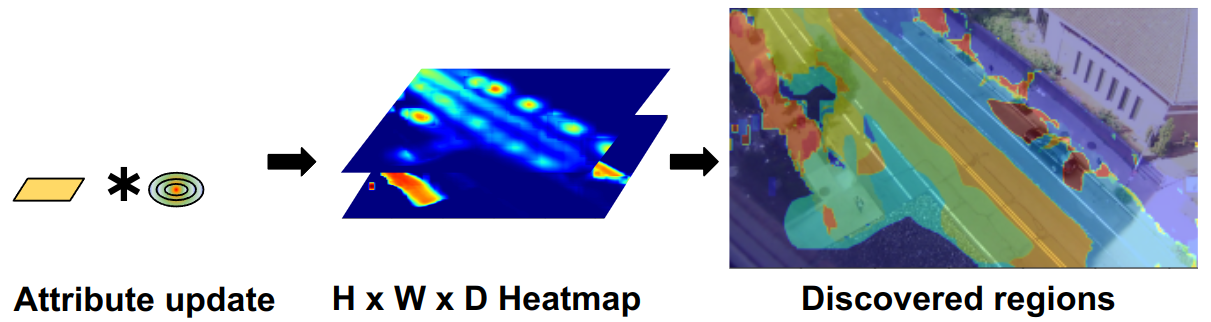}
    \caption{The region discovery process. We form a heatmap over all training objects that belong to a certain location.  After collecting the histogram for each region, we cluster and form semantic regions.}
    \label{fig:fullresupdate}
    \vspace{-0.4cm}
\end{figure}

After constructing $\vM$, we need to discover the regions that share similar activities. To capture the relationship between attribute channels, we use a Gaussian Mixture model with full covariance (Eqn.~\ref{eq:GMM}) to learn $K$ modes from $\vM$. In Section 4 we verify the importance of using full covariance compared to other alternatives. The learned modes can be used as the labels for each discovered region, as shown in Fig.~\ref{fig:fullresupdate}. 
\begin{equation}
\vspace{-0.2cm}
\begin{aligned}
p(\vf) =  \sum^{K}_{i=1}\pi_i \cdot p_i(\vf) \quad \text{with} \quad p_i(\vf)\sim \mathcal{N}(\mu_i,\Sigma_i)
\end{aligned}
\vspace{-0.2cm}
\label{eq:GMM}
\end{equation}

\subsection{Regional Model}

We use another Gaussian Mixture Model (GMM) as the basis for representing the normality distribution of the features in each region.
Different regions may exhibit more or less appearance/motion diversity in the objects that pass through them, resulting in variability in the number of components required for the GMM to adequately represent the underlying data distribution.  In our experiments, we use the Bayesian Information Criteria (BIC) score to guide the selection of the number of Gaussian modes for each region. 

In order to select the optimal $K$ in region discovery, we propose a simple evaluation metric based on symmetric KL divergence. A good region discovery method should result in regions that contain different normalcy distributions. Thus, each region's $p_i(f)$ normalcy model should assign a low likelihood to normal features from other regions. For all region pairs $i,j \in K, i \neq j$,  we use the average symmetric KL-divergence $\mathbf{\mu}_{KL}^K$  (Eqn.~\ref{eq:kl_divergence}) of sample scores to measure the region separation quality.
\begin{equation}
\begin{aligned}
    D_{\text{sym}}(P_i, P_j) &= D_{\text{KL}}(P_i \| P_j) + D_{\text{KL}}(P_j \| P_i)\\
    \mathbf{\mu}_{KL}^K&= \frac{1}{K(K-1)} \sum_{i \neq j} D_{\text{sym}}(P_i, P_j)\\
    \end{aligned}
    \label{eq:kl_divergence}
\end{equation}
In contrast to rectangular grids, in which each region may only contain a few objects' features as training samples, using our larger discovered regions drastically increases the number of training samples for each regional model.

\noindent\textbf{Learned Regional Normalcy.} As a byproduct of our algorithm, we can use the cluster results and the learned mixtures from GMMs to visualize the prototypical normal activities within each region.  For example, we can find the object in the training data whose feature is nearest to the mean of each Gaussian in a region's mixture, as in the right hand side of Figure \ref{fig:semantic_nn}.



\begin{figure*}[!t]
\begin{center}
\includegraphics[width=0.99\linewidth]{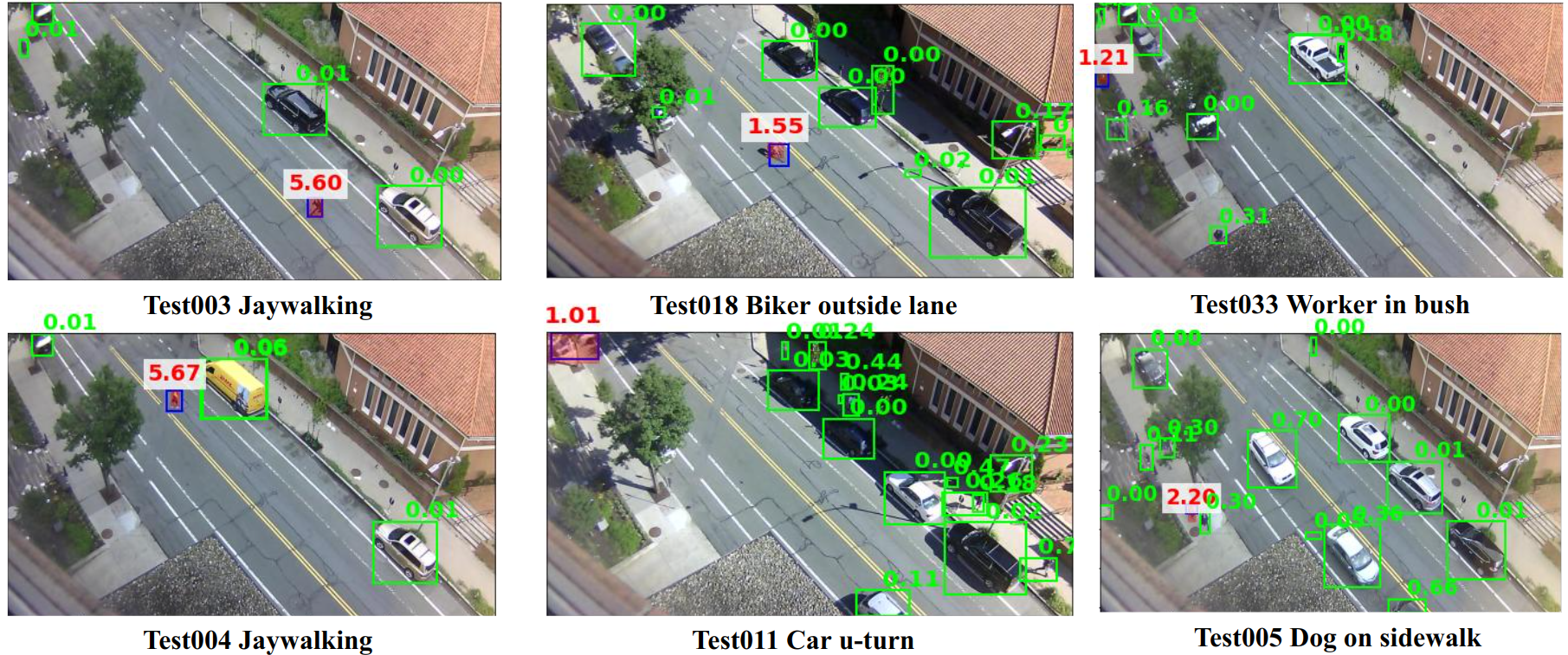}
\vspace{-0.6cm}
\end{center}
   \caption{Examples of anomaly detection results. \textcolor{green}{Green} boxes indicate normal objects, \textcolor{blue}{blue} boxes indicate ground truth anomalies, and \textcolor{red}{red} indicates detected anomalies. Anomaly scores for detected objects are shown above the bounding boxes. }
   \vspace{-0.4cm}
\label{fig:sucees_fig}
\end{figure*}
\subsection{Inference}
During inference, we extract object tracklets and features just as in the training process.  We assign objects to their enclosing region (either a rectangular box as in comparison methods or the non-rectangular regions discovered by our method) based on the first center position of their tracklets and evaluate using the corresponding model shown in Figure \ref{fig:system_overview}. For GMM models, we use the negative log-likelihood (NLL) as the indicator of anomaly. To compute a frame-level anomaly score, we simply select the highest NLL score from tracklets within each frame.


\section{Experimental Results}
\label{sec:exp}
\subsection{Dataset and Evaluation Metrics}
\noindent\textbf{Datasets.} We mainly evaluate our method on the Street Scene dataset \cite{ramachandra_street_2020} since it is one of the few publicly available datasets to address spatial-context anomalies.  This contains 46 training and 35 testing videos under a single-scene setting with 720$\times$1280 resolution.  The training and testing videos contain 56,847 and 146,410 frames respectively. The normal videos contain typical events like pedestrians walking and vehicles/cyclists moving properly in their lanes. However, the testing videos contain 17 types of spatial-context anomalies such as jaywalking, cars outside the correct lane, and cyclists on the sidewalk. We also report our performance on the widely adopted Ped2 \cite{weixin_li_anomaly_2014} and ShanghaiTech \cite{liu_future_2018} datasets. Since ShanghaiTech contains 13 different scenes, our method needs to create region maps for each camera and train the models separately. 

\noindent\textbf{Evaluation criteria.} We use both the standard frame level Area-Under-Curve (AUC) \cite{weixin_li_anomaly_2014} and the Region/Track-based Detection Criteria (RBDC/TBDC) proposed in \cite{ramachandra_street_2020}. For detailed definition, we refer readers to \cite{ramachandra_survey_2020}. We report Micro-AUC as the frame level evaluation metric that concatenates all test videos and computes the AUC score.
RBDC and TBDC measure the AUC of the region/track matching rate across false positive rates ranging from 0 to 1. While \cite{ramachandra_street_2020} highlights the shortcomings of using the frame-level AUC metric (e.g., its inability to account for multiple false positives or multiple anomalies within a frame), RBDC and TBDC also have limitations. Most studies that employ RBDC/TBDC metrics in their evaluations adhere to the original thresholds: a region match is defined as having at least 0.1 intersection-over-union between the predicted and ground truth regions, and a track match is considered successful if at least 10\% of the anomalous trajectory is detected. 

\subsection{Implementation details}
We provide implementation details in this section, for more detailed information, please refer to supplementary.

\begin{figure*}[htbp!]
    \centering
    \includegraphics[width=1\linewidth]{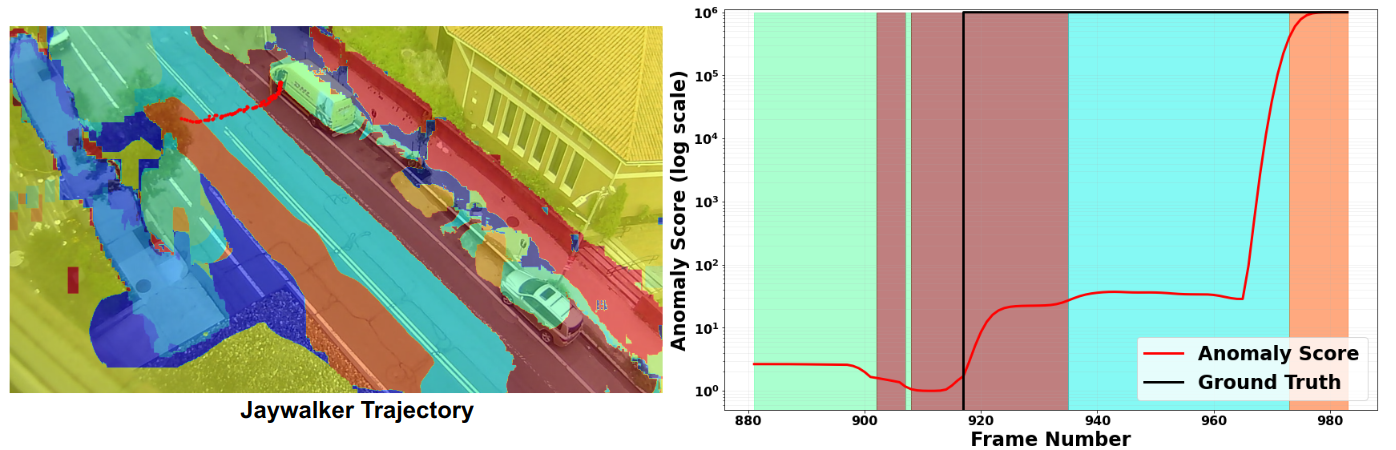}
    \vspace{-0.7cm}
    \caption{An example of the anomaly detection response to a jaywalker trajectory in \textbf{Test004}. The jaywalker exits a car to the bike lane and walks across the traffic lane.
    \textbf{Left}: The \textcolor{red}{red} trajectory indicates the jaywalker trajectory captured by our method. \textbf{Right}:  The anomaly detection scores for this object trajectory. The background colors of the graph correspond to distinct spatial regions automatically obtained from region discovery. The jaywalker motion is normal in the parking region and bike lane (light green/maroon backgrounds) but our method generates a high anomaly score when the jaywalker is in the traffic lane regions (light blue and orange backgrounds). }
    \label{fig:temporal_region}
    \vspace{-0.4cm}
\end{figure*}

\noindent\textbf{Object detection fusion.} The detection results are created from both a pre-trained YOLOv8-x \cite{yolov8} model and background subtraction. To obtain foreground objects, we initialize a Mixture of Gaussians background model \cite{stauffer1999adaptive} with variance threshold set to 16 at every individual video clip.

\noindent\textbf{Acquiring tracklets.} We set the DeepSORT algorithm with a minimum of 3 frames associated for each track. $t_w$ is set to 3 for Street Scene and ShanghaiTech and 1 for UCSD Ped2.   \\
\noindent\textbf{Region discovery and normalcy model settings.} 
The GMMs for region discovery and for normalcy modelling are set to use full covariance. The maximum component per region of BIC search is set to 20. We search the parameter $K$ over the range [2, 16]. We apply Gaussian filtering to smooth the anomaly score. 

\subsection{Quantitative Analysis}

Table \ref{tab:SS_bench} summarizes our algorithm's performance on the context-dependent Street Scene dataset.  We use a 9$\times$16 grid (i.e., 80$\times$80 pixel regions) as a baseline, denoted ``Grid Sep.'' in the table.  Our method outperforms the previous algorithms in terms of the frame-level AUC and the RBDC criterion. While other methods achieve higher TBDC, our algorithm demonstrates competitive TBDC with 1-2 orders of magnitude fewer regions and compact  feature representation. This indicates that our feature design can enable consistent detection for each anomaly trajectory.

\begin{table}[htbp!]
\centering
\resizebox{\linewidth}{!}{
\begin{tabular}{|c c c c c|}
\hline
\textbf{Method} & \textbf{Reg.}/\textbf{Dim.} & \textbf{RBDC}& \textbf{TBDC} & \textbf{AUC}  \\
\hline
Flow\cite{ramachandra_street_2020} & 2304/11200&11.0&52.0 & 51.0 \\
FG\cite{ramachandra_street_2020}  & 2304/22400& 21.0& 53.0 & 61.0 \\
EVAL\cite{singh_eval_2022} & 900/513& 24.3 & \underline{64.5} & - \\
T-EVAL\cite{singh_tracklet-based_2024} & - /150& \underline{30.9} & \textbf{72.9} & - \\
\hline
Grid Sep.  & 144/\textbf{6}& 25.9 & 56.1 & 70.2\\
Region Disc.  &\textbf{12}/\textbf{6}& \textbf{34.0} & 62.5 & 67.0\\
\hline
\end{tabular}
}
\caption{Performance comparison on the Street Scene Dataset; Bold indicates the best performance and underline indicates the second best performance. Reg.~and Dim.~stand for the number of regions and dimension of feature.  }
\label{tab:SS_bench}
\vspace{-0.2cm}
\end{table}


\begin{table}[!h]
\begin{tabular}{|c c c c c|}
\hline
\textbf{Method} & \textbf{Venue} & \textbf{RBDC}& \textbf{TBDC} & \textbf{AUC}  \\

\hline
AED \cite{georgescu_background-agnostic_2021} & \textit{PAMI21} &69.2 &93.2&98.7 \\
HF2VAD \cite{liu_hybrid_2021} & \textit{ICCV21}&- &- & \textbf{99.3}\\
SSMTL \cite{georgescu_anomaly_2021} &\textit{CVPR21} &72.8& 91.2 & 97.5\\
EVAL \cite{singh_eval_2022} &\textit{CVPR22} &\textbf{87.4}& 95.1 & -\\
HSC \cite{sun_hierarchical_2023}& \textit{CVPR23} & -&- & 98.2 \\
T-EVAL \cite{singh_tracklet-based_2024} &\textit{CVPRW24} &84.2& \underline{97.9} & -\\
\hline
\textbf{Ours} & - & \underline{84.8} & \textbf{99.1} & \underline{99.0} \\
\hline
\end{tabular}
\caption{Performance comparison on the UCSD Ped2 Dataset.}
\label{tab:ped2_perf}
\end{table}

Although the main focus of our proposed method is to detect spatial-context dependent anomalies, we also test our algorithm on the UCSD Ped2 dataset \cite{weixin_li_anomaly_2014} (Table \ref{tab:ped2_perf}) and the ShanghaiTech dataset \cite{liu_future_2018} (Table \ref{tab:STC_perf}).  Our method uses a simple 6-dimensional feature to achieve state-of-the-art performance in video anomaly tracking and second best video anomaly localization in UCSD Ped2,  further illustrating the effectiveness and compactness of our proposed method. Our model also shows competitive performance on the ShanghaiTech dataset. 
\begin{table}[!h]
\begin{tabular}{|c c c c c|}
\hline
\textbf{Method} & \textbf{Venue} & \textbf{RBDC}& \textbf{TBDC} & \textbf{AUC}  \\

\hline
AED \cite{georgescu_background-agnostic_2021} & \textit{PAMI21} &41.3 &78.8&82.7\\
HF2VAD \cite{liu_hybrid_2021} & \textit{ICCV21}&- &- & 76.1\\
SSMTL \cite{georgescu_anomaly_2021} &\textit{CVPR21} &42.8& 83.9 & \textbf{83.4}\\
EVAL \cite{singh_eval_2022} &\textit{CVPR22} &\underline{59.2} & \textbf{89.4}  &76.6 \\
HSC \cite{sun_hierarchical_2023}& \textit{CVPR23} & -&- & \textbf{83.4} \\
T-EVAL \cite{singh_tracklet-based_2024} &\textit{CVPRW24} &\textbf{59.6}&87.6 & -\\
\hline
\textbf{Ours} & - & 46.8 & 73.2 & 81.3\\
\hline
\end{tabular}
\caption{Performance comparison on the ShanghaiTech Dataset.}
\label{tab:STC_perf}
\vspace{-0.4cm}
\end{table}

\subsection{Qualitative examples}

Figure \ref{fig:sucees_fig} illustrates successful anomaly detection results from our algorithm on the Street Scene dataset, demonstrating its ability to detect difficult examples like a parking attendant ticketing a car and a dog on the sidewalk.  The same anomaly threshold is used in all examples of the figure.

We drill down into the anomaly detection criterion in Figure \ref{fig:temporal_region}, illustrating the response as a pedestrian exits a car into the bike lane and later walks across the traffic lanes. Within the bike lane and parking region (maroon), the anomaly score depends on the jaywalker's motion attributes. When the jaywalker appears in the traffic lanes, our method immediately generates a high anomaly score.

\subsection{Parameter analysis}
\label{sec:exp}

In this section, we validate our choices for important components and parameters in our proposed method. \\
\noindent\textbf{Region proposal vs.~semantic segmentation.}
To evaluate the effectiveness of our region proposal, we directly compare the performance in the Street Scene dataset with region maps generated from our region proposal and the off-the-shelf semantic segmentation model SegFormer \cite{xie2021segformer} (with the largest variant B5 pretrained on ADE20K \cite{zhou_scene_2017}) shown in Table \ref{tab:seg_comp}. We use the average logit output of all training videos with blocking foreground objects to generate the region map (SegFormer) of a single scene. We also mimic the process of our region proposals to cluster the output logits into the same number of clusters (SegFormer clustering). Since the segmentation model lacks motion cues, there is a significant performance improvement of our region proposals over the off-the-shelf semantic segmentation model. For more details, see the supplementary material.  

\begin{table}[htbp!]
    \centering
    \resizebox{\linewidth}{!}{
    \begin{tabular}{c |ccc}
    \hline
       \textbf{Region map}  & \textbf{RBDC}& \textbf{TBDC} & \textbf{AUC}\\
      \hline
       SegFormer & 20.5 &51.5&59.3\\
     SegFormer clustering    & 29.1 &46.7&61.3\\
    \textbf{Ours}    & \textbf{34.0} &\textbf{62.5}&\textbf{67.0}\\  
      \hline
    \end{tabular}
    }
    \caption{Performance using different region proposals in the Street Scene Dataset.}
    \label{tab:seg_comp}
    \vspace{-0.2cm}
\end{table}
\noindent\textbf{Deep features vs.~object class for spatial context-dependent anomalies.} As we mentioned above, using object classifications instead of deep appearance features can significantly reduce computational load and the number of model parameters. Table \ref{tab:clip_perf} compares the performance using full CLIP features \cite{radford_learning_2021} vs.~the one-hot classification of expected objects in the scene.  We see that the performance is actually very similar. 
\begin{table}
\vspace{-0.2cm}
\begin{center}
    \begin{tabular}{|c|c|c|c|}
\hline
\textbf{Feature type} & \textbf{Feature Dim.} & \textbf{RBDC}& \textbf{TBDC}   \\
\hline

CLIP &512& 33.1 & 60.1\\

Object CLS &\textbf{4}& \textbf{34.0} & \textbf{62.5} \\
\hline
\end{tabular}
\caption{Performance comparison between using CLIP features and object classification in the Street Scene Dataset.}
\label{tab:clip_perf}
\end{center}
\vspace{-0.4cm}
\end{table}

\noindent\textbf{Optimal region discovery for spatial context-dependent anomalies.} The performances for different numbers of regions are summarized in  Table \ref{tab:num_regions}, showing that 12 region model using KL-divergence criteria $\mu_{KL}$ achieves both the highest divergence and best performance in our experiments. We also report the divergence in the grid separation scenario; the low divergence further indicates the redundancy of having overly granular regions.

\begin{table}[htbp!]
    \centering
    \resizebox{\linewidth}{!}{
    \begin{tabular}{|c|c|c|c|c|}
    \hline
       \textbf{Number of Regions} &$\boldsymbol{\mu}_{KL}$ &\textbf{RBDC} & \textbf{TBDC} & \textbf{AUC}\\
      \hline
      6 & 9.23&21.2 &48.9&61.0\\
       8 &  10.21&25.9 &50.9&64.0\\
        10 & 11.64& 28.9 &53.9&66.0\\
        12 &\textbf{14.75}& \textbf{34.0} &\textbf{62.5}&67.0\\
        16& 13.47 & 29.9 &57.9&67.0\\
     144 & 4.99& 25.9 &56.1 &\textbf{70.2} \\ 
      \hline
    \end{tabular}
        }
    \caption{Performance using different numbers of discovered regions in Street Scene Dataset. }
    \label{tab:num_regions}
    \vspace{-0.4cm}
\end{table}


To verify the effectiveness of different attribute features, we also performed an ablation study in Table \ref{tab:ablation_branch}. We can see that all attributes are required for the best performance, and we observe that the different attributes often complement each other.  

\begin{table}[htbp!]
    \centering
    \resizebox{\linewidth}{!}{
    \begin{tabular}{|c c c|c|c|c|}
    \hline
       \textbf{Obj. Cls.} & \textbf{Ori.}& \textbf{Mag.} & \textbf{RBDC}& \textbf{TBDC} & \textbf{AUC}\\
      \hline
       \checkmark&  & & 31.2 &45.1&67.9\\
        &\checkmark  & & 29.0 &49.8&63.3\\
        & &\checkmark  &  10.7 &52.4&62.0\\
        \checkmark&  &\checkmark & 26.2 &48.0&66.0\\
        \checkmark&\checkmark & & \textbf{35.6} &56.7&\textbf{73.1}\\
     \checkmark&\checkmark &\checkmark & 34.0 &\textbf{62.5}&67.0\\
     
      \hline
    \end{tabular}
    }
    \caption{Performance using different feature branches in Street Scene dataset. Obj., Cls., Ori. and Mag. stand for object, class, motion orientation, and motion magnitude respectively. \rnote{} }
    \label{tab:ablation_branch}
    \vspace{-0.4cm}
\end{table}

\section{Conclusions}
\begin{figure}[htbp!]
    \centering
    \vspace{-0.3cm}
    \includegraphics[width=1\linewidth]{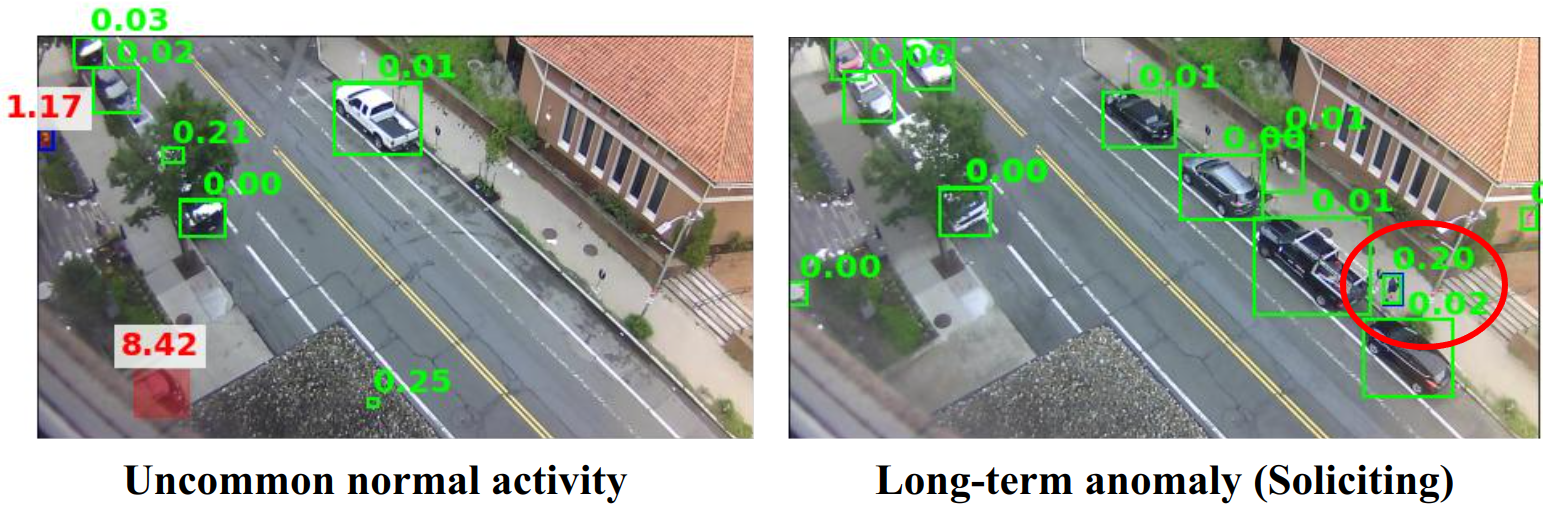}
    \caption{Examples of false alarms and misses. \textcolor{green}{Green} boxes indicate normal objects, \textcolor{blue}{blue} boxes indicate ground truth anomalies, and \textcolor{red}{red} indicates detected anomalies. }
    \label{fig:failure_fig}
\vspace{-.6cm}
\end{figure}

We proposed a simple but effective method for detecting spatial-context anomalies using region discovery. There are several limitations of the proposed method as shown in Figure \ref{fig:failure_fig}. Since our model relies on appearance and motion features within a short time window, its ability to detect long-term trajectory-oriented anomalies like soliciting is limited. This also partially explains why we achieve state-of-the-art performance on the region-based criteria RBDC but not on the track-based criteria TBDC. We identified several opportunities to improve the proposed algorithm in future work. First, the current hard region assignment might magnify the impact of incorrect region clustering. It would be helpful to explore soft region assignment which should alleviate this problem. For more person action oriented datasets like ShanghaiTech, incorporating pose features \cite{reiss_attribute-based_2022,hirschorn_normalizing_2023} could further boost performance.

\section{Acknowledgement}
This material is based upon work supported by the U.S. Department of Homeland Security
under Grant Award 22STESE00001-04-00. The views and conclusions contained in this
document are those of the authors and should not be interpreted as necessarily representing
the official policies, either expressed or implied, of the U.S. Department of Homeland
Security.
{
    \small
    \bibliographystyle{ieee_fullname}
    \bibliography{submitbib}
}

\end{document}